\documentclass[letterpaper]{article} 
\usepackage{aaai23}  
\usepackage{times}  
\usepackage{helvet}  
\usepackage{courier}  
\usepackage[hyphens]{url}  
\usepackage{graphicx} 
\usepackage{natbib}  
\usepackage{caption} 
\usepackage{algorithm}
\usepackage{algorithmic}

\usepackage{newfloat}
\usepackage{listings}
\DeclareCaptionStyle{ruled}{labelfont=normalfont,labelsep=colon,strut=off} 
\floatstyle{ruled}
\newfloat{listing}{tb}{lst}{}
\floatname{listing}{Listing}
\usepackage{makecell}
\usepackage{booktabs}  
\usepackage{threeparttable}
\usepackage{multirow}
\usepackage{epstopdf}
\usepackage{algorithm}
\usepackage{algorithmic}
\usepackage{amsfonts} 
\usepackage{epstopdf}
\usepackage{booktabs}
\usepackage{epsfig} 
\usepackage{subfigure}
\usepackage{graphicx}
\usepackage{amsmath}
\usepackage{amssymb, amsthm}
\usepackage{booktabs}

\newtheorem{theorem}{Theorem}
\newtheorem{remark}{Remark}

\usepackage[table]{xcolor}
\usepackage{wrapfig}
\definecolor{mygray}{gray}{.9}
\usepackage{colortbl}

\title{Unlabeled Imperfect Demonstrations in Adversarial Imitation Learning}
\author{
    Yunke Wang\textsuperscript{\rm 1}, Bo Du\textsuperscript{\rm 1}\footnote{Corresponding author}, Chang Xu\textsuperscript{\rm 2}$^\ast$
    \\
}
\affiliations{
    \textsuperscript{\rm 1}School of Computer Science, National Engineering Research Center for Multimedia Software, \\Institute of Artificial Intelligence, and Hubei Key Laboratory of Multimedia and Network Communication Engineering, Wuhan University, Wuhan, China.\\
    \textsuperscript{\rm 2}School of Computer Science, Faculty of Engineering, The University of Sydney, Australia.\\
    \{yunke.wang, dubo\}@whu.edu.cn, c.xu@sydney.edu.au
}

\usepackage{bibentry}

\begin{document}

\maketitle

\begin{abstract}
Adversarial imitation learning has become a widely used imitation learning framework. The discriminator is often trained by taking expert demonstrations and policy trajectories as examples respectively from two categories (positive vs. negative) and the policy is then expected to produce trajectories that are indistinguishable from the expert demonstrations. But in the real world, the collected expert demonstrations are more likely to be imperfect, where only an unknown fraction of the demonstrations are optimal. Instead of treating imperfect expert demonstrations as absolutely positive or negative, we investigate \textit{unlabeled} imperfect expert demonstrations as they are. A positive-unlabeled adversarial imitation learning algorithm is developed to dynamically sample expert demonstrations that can well match the trajectories from the constantly optimized agent policy. The trajectories of an initial agent policy could be closer to those non-optimal expert demonstrations, but within the framework of adversarial imitation learning, agent policy will be optimized to cheat the discriminator and produce trajectories that are similar to those optimal expert demonstrations. Theoretical analysis shows that our method learns from the imperfect demonstrations via a self-paced way. Experimental results on MuJoCo and RoboSuite platforms demonstrate the effectiveness of our method from different aspects.
\end{abstract}

\section{Introduction}
Reinforcement Learning (RL) \cite{sutton2018reinforcement,kaelbling1996reinforcement} provides an effective framework for solving sequential decision-making problems \cite{silver2016mastering,van2016deep,zha2021douzero}. It aims to learn a good policy by rewarding the agent's action during its interaction with the environment. A well-formulated reward can recover the best policy, yet this complex reward engineering \cite{amodei2016concrete} in real-world tasks can make RL fail sometimes. By contrast, it could be more practical to introduce imitation learning (IL) \cite{hussein2017imitation,zheng2022imitation}: a popular learning paradigm to guide policy learning by directly mimicking expert behaviors. A basic approach of IL is Behavioral Cloning (BC) \cite{pomerleau1988alvinn}, in which the agent observes the action of the expert and learns a mapping from state to action via regression. However, this offline training manner may suffer from compounding errors \cite{brantley2019disagreement, xu2020error,tu2022sample} when the agent executes the policy, leading it to drift to new and dangerous states. Instead, Adversarial Imitation Learning (AIL) encourages the agent to cover the distribution of the expert policy, which can result in a more precise policy. Generative Adversarial Imitation Learning (GAIL) \cite{ho2016generative} is the most prominent work of AIL and it inherits the framework of Generative Adversarial Nets (GAN) \cite{goodfellow2014generative}. After GAIL, there are many variants \cite{li2017infogail, fu2018learning,peng2018variational,dadashi2020primal,cai2021seeing} to further enhance the performance of AIL from various aspects.

Existing imitation learning methods achieve promising results under the assumption that the given expert demonstrations are of high quality \cite{hussein2017imitation}. However, there is a fact that most of them would fail when injecting some non-optimal demonstrations into expert demonstrations, which results in the imperfect demonstrations issue in IL \cite{wu2019imitation, ross2011reduction}. This issue is of practice since it could be costly to collect purely optimal demonstrations in the real world. Therefore, how to learn a good policy from a mixture of optimal and non-optimal demonstrations is crucial to bridging the applicable gap of IL from the simulator to real-world tasks.

Confidence-based methods are popular and effective to address imperfect demonstrations issue in imitation learning. The key lies in how to acquire proper confidence for each expert demonstration. In 2IWIL \cite{wu2019imitation} and IC-GAIL \cite{wu2019imitation}, an annotator is employed to manually label confidence for a fraction of demonstrations. 
The former is a two-stage method, which predicts the confidence for the remaining unlabeled demonstrations first and then conducts a weighted imitation learning framework. The latter combines these two steps in a single objective function instead. 
WGAIL \cite{wang2021learning} successfully connects confidence estimation to the discriminator in GAIL, and BCND \cite{sasaki2021behavioral} demonstrates confidence can be derived by the agent policy itself. Therefore, these two methods relax the assumption on the labeled confidence and can be conducted without exposure to prior information. 
The confidence estimation in these two methods largely relies on the model's training status itself, but there might be some extreme situations where the model collapses and fails to predict informative confidence. For example, the high ratio of contamination in expert demonstrations could seriously hurt the training process of IL, which can further lead to the collapse of confidence estimation. Confidence-based methods under such cases might be hard to even outperform their baseline.

Instead of estimating the precise confidence, our thought is to adopt a better training scheme for adversarial imitation learning with imperfect demonstrations via positive-unlabeled learning. This results in our method UID, which is general and can be equipped with various adversarial imitation learning backbones. Specifically, the imperfect demonstrations in UID are treated as unlabeled data, in which there exists a fraction of demonstrations that can well match the agent demonstrations. The positive-unlabeled adversarial imitation learning process can therefore be formulated by dynamically sampling demonstrations that resemble the behavior of the constantly optimized agent policy. The agent policy might produce demonstrations similar to the non-optimal demonstrations at the early training stage, yet it will be optimized to cheat the discriminator and produces demonstrations resembling those optimal demonstrations within the framework of adversarial training. Theoretical analysis shows UID gradually makes the agent cover more samples in unlabeled demonstrations via a self-paced way. Experimental results in MuJoCo \cite{todorov2012mujoco} and RoboSuite \cite{fan2018surreal} demonstrate the effectiveness of UID from different aspects.  

\section{Related Work}
In this section, we briefly review the existing researches on imitation learning with imperfect demonstrations. We roughly divided them into two categories, \textit{i.e.}, confidence-based methods and preference-based methods.

\subsubsection{Confidence-based methods}
Instance reweighting has been widely used in various machine learning problems \cite{zhang2020geometry, ren2020not, zhong2021learning, qiu2022dynamic} and gains great success. 2IWIL \cite{wu2019imitation} and IC-GAIL \cite{wu2019imitation} first investigate the capacity of the weighting scheme in imitation learning and find it effective in dealing with imperfect demonstrations. However, the assumption that a fraction of demonstrations should be manually labeled with confidence is a strong prior and hard to satisfy in the real world. Additionally, different human annotators may have different judgments on the goodness of demonstrations. 
The following works \cite{wang2021learning, zhang2021confidence, ijcai2021-0434, chen2022anomaly} thus focus on how to relax the assumption when estimating the confidence. To name a few, CAIL \cite{zhang2021confidence} considers to introduce a small fraction of ranked trajectories to help with the confidence estimation during the training. WGAIL \cite{wang2021learning} proves that the optimal confidence should be proportional to the exponential advantage function, and then connects advantage with agent policy and the discriminator in GAIL. An alternating interaction between weight estimation and GAIL training therefore holds. 
There are also some researches \cite{sasaki2021behavioral, kim2021demodice, xu2022discriminator, liu2022robust} on addressing imperfect demonstrations issue in offline imitation learning. BCND \cite{sasaki2021behavioral} is a weighted behavioral cloning method, with action distribution of learned policy as confidence. However, when sub-optimal demonstrations occupy the major mode within imperfect demonstrations, the confidence distribution is likely to drift to the sub-optimal demonstrations and assign higher confidence to them. 
DemoDICE \cite{kim2021demodice} performs offline imitation learning with a KL constraint between the learned policy and supplementary imperfect demonstrations to efficiently utilize additional demonstration data.

\subsubsection{Preference-based methods}
Preference-based methods have been proved to be effective in policy learning. \cite{christiano2017deep} firstly applied active preference learning to Atari games, asking the expert to select the best of two trajectories generated from an ensemble of policies. The policy is learned to maximize the reward defined by expert preference during the interaction. T-REX \cite{brown2019extrapolating} aims to extrapolate a reward function by ranked trajectories. The learned reward function can well explain the rankings, and thus is informative to be used as feedback for the agent. T-REX only requires precise rankings of trajectories, yet does not set constraints on data quality. It can thus perform quite well even with no optimal trajectories. D-REX \cite{brown2020better} further relaxes T-REX's constraint on rankings. It learns a pre-trained policy by behavioral cloning first, and the ranked trajectories can be generated by injecting different noises into its action. SSRR \cite{chen2021learning} fixes the possible error of rankings by defining a new structure of the reward function. 

\section{Preliminary}
In this section, we briefly review the definition of Markov Decision Process (MDP) and adversarial imitation learning.
 
\subsubsection{Markov Decision Process (MDP)}
MDP is popular to formulate the reinforcement learning (RL) \cite{DBLP:books/wi/Puterman94} and imitation learning (IL) problems. An MDP normally consists six basic elements $\mathcal{M}=(\mathcal{S}, \mathcal{A}, \mathcal{P}, \mathcal{R}, \gamma, \mu_0)$, where $\mathcal{S}$ is a set of states, $\mathcal{A}$ is a set of actions, $\mathcal{P}:\mathcal{S}\times\mathcal{A}\times\mathcal{S}\rightarrow [0, 1]$ is the stochastic transition probability from current state $s$ to the next state $s'$, $\mathcal{R}:\mathcal{S}\times\mathcal{A}\rightarrow \mathbb{R}$ is the obtained reward of agent when taking action $a$ in a certain state $s$, $\gamma \in [0,1]$ is the discounted rate and $\mu_0$ denotes the initial state distribution. Given an trajectory $\tau  = \{(s_t, a_t)\}_{t=0}^{T}$, the return $R(\tau)$ is defined as the discounted sum of rewards obtained by the agent over all episodes, , $R(\tau)=\sum_{t=0}^{T} \gamma^k r(s_k, a_k)$ and $T$ is the number of steps to reach an absorbing state. The goal of RL is thus to learn a policy that can maximize the expected return over all episodes during interaction. For any policy $\pi:\mathcal{S}\rightarrow\mathcal{A}$, there is an one-to-one correspondence between $\pi$ and its occupancy measure $\rho_\pi:\mathcal{S}\times\mathcal{A}\rightarrow[0, 1]$.

\subsubsection{Adversarial Imitation Learning (AIL)} 
Adversarial imitation learning addresses IL problems from the perspective of distribution matching. By minimizing the distance between distributions of agent demonstrations and expert behaviors, AIL can thus recover the expert policy. 
Generative Adversarial Imitation Learning (GAIL) \cite{ho2016generative} is the most representative work of AIL, which directly applies the general GAN framework \cite{goodfellow2014generative} into adversarial imitation learning. Given a set of expert demonstrations $\mathcal{D}_e$ drawn from the expert policy $\pi_e$, GAIL aims to learn an agent policy $\pi_\theta$ by minimizing the Jensen-Shannon divergence between $\rho_{\pi_\theta}$ and $\rho_{\pi_e}$ \cite{ke2020imitation}. In the implementation, a discriminator is introduced to distinguish demonstrations from expert policy and agent policy, yet the agent policy tries its best to `fool' the discriminator. This results in a minimax adversarial objective as follow,
\begin{align}
    \min_\theta \max_\psi \;& \mathbb{E}_{(s,a)\sim\rho_{\pi_e}} [\log D_\psi(s,a)] \\ \nonumber & +\mathbb{E}_{(s,a)\sim\rho_{\pi_\theta}} [\log(1-D_\psi(s,a))].
    \label{gail}
\end{align}
The agent is trained to minimize the outer objective function $\mathbb{E}_{(s,a)\sim\rho_{\pi_\theta}} [\log(1-D_\psi(s,a))]$, and therefore the output of $-\log(1-D_\psi(s,a))$ can be regarded as reward. Regular RL methods like TRPO \cite{schulman2015trust}, PPO \cite{schulman2017proximal} and SAC \cite{haarnoja2018soft} can be thus used to update the agent policy $\pi_\theta$. 

\section{Methodology}
Most adversarial imitation learning methods achieve promising results in benchmark tasks with a non-trivial assumption that the given expert demonstrations should be optimal. However, querying the expert for a large amount of optimal behaviors can be expensive in some real-world tasks. By contrast, it could be more realistic to collect mixed demonstrations with only a fraction of optimal samples. In this paper, we consider this practical setting and investigate how to ensure a promising performance when dealing with imperfect demonstrations. 

\subsection{Proposed Method: UID}
In our setting, we have a mixture set of expert demonstrations $\mathcal{D}_e$ that contains both optimal demonstrations and non-optimal demonstrations. Since the specific information about the demonstrations' optimality is unknown, we consider to regard $\mathcal{D}_e$ as \textit{unlabeled} demonstrations and label their categories dynamically based on the status of agent policy. 
Supposing $\rho_{\pi_{\widehat{\theta}}}$ represents the distribution of a fraction of unlabeled demonstrations that can well match agent demonstrations from $\rho_{\pi_\theta}$, 
we model $\rho_{\pi_e}$ as a mixture of distributions
\begin{align}
    \rho_{\pi_e}(s,a) = (1-\alpha) \rho_{\pi_\epsilon} (s,a) + \alpha \rho_{\pi_{\widehat{\theta}}} (s,a),
\end{align}
where $\alpha \in [0,1]$ is the mixing proportion of the matched distribution $\rho_{\pi_{\widehat{\theta}}}$, and $\rho_{\pi_\epsilon}$ can be regarded as the distribution of those remaining demonstrations in $\mathcal{D}_e$. We denote $\pi_\epsilon$ as the \textit{residual} policy.
In plain adversarial imitation learning, all unlabeled demonstrations are simply labeled as positives in discriminator training. However, this training scheme only makes sense when the labeled demonstrations are clean. When there exists some non-optimal demonstrations, the discriminator would equally treat both optimal demonstrations and non-optimal demonstrations. Hence, agent demonstrations that resemble those imperfect data would also be assigned with high reward, which results in sub-optimal agent behavior.

Our thought is to build an arbitrary discriminator $g:(s,a)\mapsto \mathbb{R}$ that has better discriminative ability among unlabeled demonstrations $\mathcal{D}_e$. Supposing the surrogate loss function $\phi:{\mathbb{R}\times \{\pm{1}\}}\mapsto \mathbb{R}$ is a margin-based loss function for binary classification, the expectation of risk of discriminator $g$ can be expressed as
\begin{align}
    R_{\pi_e}(g) 
    =\; &(1-\alpha) \mathbb{E}_{(s, a)\sim \rho_{\pi_\epsilon}} [\phi(g(s,a))]\\\nonumber &+ \alpha \mathbb{E}_{(s, a)\sim \rho_{\pi_{\widehat{\theta}}}} [-\phi(g(s,a))].
\end{align}
The residual policy $\pi_\epsilon$ is inaccessible in our setting, however, we have $\pi_{\widehat{\theta}}$ that is assumed to be approximating the agent policy $\pi_\theta$. Therefore, we consider to replace $(1-\alpha)\rho_{\pi_\epsilon}$ with $(\rho_{\pi_e}-\alpha\rho_{\pi_{\widehat{\theta}}})$ and introduce the agent policy $\pi_\theta$ as an estimation of ${\pi_{\widehat{\theta}}}$. Then, the expected risk $R_{\pi_e}$ can be estimated by $\pi_\theta$ and $\pi_e$, and the optimal discriminator $g$ can be obtained by minimizing $R_{\pi_e}(g)$,
\begin{align}
    \label{eq4}
    \min_g R_{\pi_e}(g) = \mathcal{T} & \{0, \mathbb{E}_{(s, a)\sim \rho_{\pi_e}} [\phi(g(s,a))]\\\nonumber & - \alpha \mathbb{E}_{(s, a)\sim \rho_{\pi_\theta}} [\phi(g(s,a))] \}
    \\ \nonumber &
    + \alpha \mathbb{E}_{(s, a)\sim \rho_{\pi_\theta}} [-\phi(g(s,a))],
\end{align}
where $\mathcal{T}\{\cdot\}$ is a flexible constraint, which makes the replacement $\mathbb{E}_{(s, a)\sim \rho_{\pi_e}} [\phi(g(s,a))] - \alpha \mathbb{E}_{(s, a)\sim \rho_{\pi_\theta}} [\phi(g(s,a))]$ have the same sign as the original loss function $\mathbb{E}_{(s, a)\sim \rho_{\pi_\epsilon}} [\phi(g(s,a))]$. Eq. (\ref{eq4}) is an unbiased and consistent risk estimator of the true risk w.r.t all popular loss functions as mentioned in \cite{niu2016theoretical}. 

Considering the agent policy should also learn from this discriminator, it should be trained to produce trajectories that can `fool' the judgment of the discriminator. We therefore set up an adversarial game between $\pi_\theta$ and $g$, and obtain the following objective function $\mathcal{J}(\theta,g)$,
\begin{align}
    \label{eq5}
    \max_\theta \min_g \mathcal{J}(g, \theta)=\mathcal{T} & \{0, \mathbb{E}_{(s, a)\sim \rho_{\pi_e}} [\phi(g(s,a))]
    \\\nonumber & - \alpha \mathbb{E}_{(s, a)\sim \rho_{\pi_\theta}} [\phi(g(s,a))] \} 
    \\\nonumber & +  \alpha \mathbb{E}_{(s, a)\sim \rho_{\pi_\theta}} [-\phi(g(s,a))].
\end{align}
Eq. (\ref{eq5}) is a general objective function with an unspecific loss function $\phi$. 
However, since adversarial imitation learning methods are not always directly linked to a certain surrogate loss function, it is hard to straightly recover various AIL baselines by specifying a $\phi(g)$. By contrast, most adversarial imitation learning methods can be viewed as minimizing the different distances between occupancy measures of agent policy and expert policy. We therefore consider to connect margin-based loss function $\phi(g)$ with $f$-divergence and then write the general form of UID for various adversarial imitation learning methods. We summarize this process in the following theorem.
\begin{theorem}
For any margin-based surrogate convex loss $\phi:{\mathbb{R}\times \{\pm{1}\}}\mapsto \mathbb{R}$ in Eq. (\ref{eq4}), there is a related $f$-divergence $I_f$ such that 
\begin{align}
 \min_g R_{\pi_e}(g, \phi)=-I_f(\mu, \nu)= - \int_{s,a} \mu(s,a) f(\frac{\mu(s,a)}{\nu(s,a)}) dsda   ,
\end{align}
 where $\mu=\rho_{\pi_e}-\alpha\rho_{\pi_\theta}$, $\nu=\alpha\rho_{\pi_\theta}$ and $f:[0, \infty]\rightarrow \mathbb{R} \cup \{\infty \}$ is a continuous convex function. Then, by using variational approximation of $f$-divergence, $\min_g R_{\pi_e}(g)$ can be further written as
\begin{align}
   \max_T & \min\{0, \mathbb{E}_{(s,a) \sim \rho_{\pi_e}} [T(s,a)] \label{theo1equ}\\\nonumber &- \alpha \mathbb{E}_{(s,a) \sim \rho_{\pi_\theta}} [T(s,a)]\}-\alpha \mathbb{E}_{(s,a) \sim \rho_{\pi_\theta}} f^\ast[T(s,a)].
\end{align}
where $T(s,a)$ is the decision function related to $g$. Different choices of convex function $f$ can recover different objective function of UID adversarial imitation learning.
\label{theo1}
\end{theorem}
With the help of Theorem \ref{theo1}, we can now integrate the proposed method into various frameworks of AIL with different choices of $f$-divergence. This flexibility that combined with other models provides the proposed method a chance to get further improvement on existing adversarial imitation learning backbones. 

\subsubsection{UID-GAIL}
We provide a specific case by recovering GAIL, which is the most representative AIL methods. We consider to use Jensen-Shannon divergence and define $f(u)=-(u+1)\log\frac{u+1}{2}+u\log u$, $f^\ast(t)=-1-\log(1-\exp(t))$. By replacing $T(s,a)$ with $\log [D(s,a)]$, the objective function of UID can be written as,
\begin{align}
    \label{UIDgail}
    \min_\theta \max_\psi \ \mathcal{J}(\theta, \psi)=& \min\{0, \mathbb{E}_{(s,a) \sim \rho_{\pi_e}} \log[D_\psi(s,a)] \\\nonumber &- \alpha \mathbb{E}_{(s,a) \sim \rho_{\pi_\theta}} \log[D_\psi(s,a)]\} \\\nonumber & +  \alpha \mathbb{E}_{(s,a) \sim \rho_{\pi_\theta}} \log[1-D_\psi(s,a)].
\end{align}
The practical optimization of UID-GAIL is summarized in Algorithm \ref{al1}.

\subsubsection{UID-WAIL}
We also show the flexibility of UID with other popular AIL methods, \textit{i.e.}, WAIL \cite{xiao2019wasserstein}.
Recall that Theorem \ref{theo1} makes it possible to recover specific adversarial imitation learning baselines by defining different $f$-divergence functions, however, the Wasserstein distance metric used in WAIL is not strictly an $f$-divergence. Therefore, we begin with Total Variation (TV), which is a kind of $f$-divergence that is related to Wasserstein distance. The $f$ function in total variation is defined as $f(u)=\frac{1}{2}|u-1|$, therefore we have $f^\ast(t)=t$. By defining critic $r_\psi(s,a)=T(s,a)$, we then re-write Eq. (\ref{theo1equ}) as,
\begin{align}
    \max_\psi & \min\{0, \mathbb{E}_{(s,a) \sim \rho_{\pi_E}} [ r_\psi(s,a)] \label{rdtwail} \\\nonumber &
    - \alpha \mathbb{E}_{(s,a) \sim \rho_{\pi_\theta}} [ r_\psi(s,a)] \} - \alpha \mathbb{E}_{(s,a) \sim \rho_{\pi_\theta}} [ r_\psi(s,a)].
\end{align}
TV can be regarded as the Wasserstein distance with respect to 1-Lipschitz constraint on $r_\psi$. We then add this regularization on $r_\psi$ and obtain the final objective function of UID-WAIL,
\begin{align}
    \min_\theta \max_\psi \min&\big\{0, \mathbb{E}_{(s,a) \sim \rho_{\pi_e}} [ r_\psi(s,a)] - \alpha \mathbb{E}_{(s,a) \sim \rho_{\pi_\theta}} [ r_\psi(s,a)] \big\} \label{puwail} \nonumber\\ & -  \alpha \mathbb{E}_{(s,a) \sim \rho_{\pi_\theta}} [ r_\psi(s,a)] + \lambda \Psi(r_\psi), 
\end{align}
where the critic $r_\psi$ serves as the reward function and $\Psi(r_\psi)=-\mathbb{E}_{(s,a)\sim \rho_{\hat{\pi}}}(||\nabla r_\psi(\hat{s},\hat{a})||_2-1)^2$ is the regularization term to satisfy the Lipschitz constraint. 

\begin{algorithm}[!t] 
	\caption{UID-GAIL} 
	\label{alg:Framwork} 
	\begin{algorithmic}[1] 
		\REQUIRE ~~\\
		Unlabeled demonstrations $\mathcal{D}_e=\{s_i,a_i\}_{i=1}^{n} \sim \rho_{\pi_e}$; 
		\\ Total iterations $N$;
		\ENSURE ~~\\ 
		The agent policy $\pi_\theta$; The discriminator $D_\psi$;
		\STATE Initialize $D_\psi$ and $\pi_{\theta}$;
		\label{ code:fram:extract }
		\FOR{iter = 1 to N}
		\label{code:fram:add}
		\STATE Sample trajectories $\{s^\theta,a^\theta\}\sim \rho_{\pi_\theta}$,  $\{s^e,a^e\}\sim \mathcal{D}_e$;\\
		\label{code:fram:classify}
		\STATE Update $D_{\psi}$ by maximizing $\mathcal{J}(\theta, \psi)$\\
		\label{code:fram:select}
		\STATE Update $\pi_\theta$ by TRPO with reward $-\log [1-D_\psi(s,a)]$;  
		\ENDFOR 
		\label{algo1}
	\end{algorithmic}
	\label{al1}
\end{algorithm}

\subsection{Theoretical Results of UID}
Since $\rho_{\pi_{\widehat{\theta}}}$ dynamically samples from $\rho_{\pi_e}$ to approximate $\rho_{\pi_\theta}$ during training, the PU discriminator will thus make the agent produce demonstrations that resemble the residual policy $\pi_\epsilon$. As $\pi_\epsilon$ changes during training as well, the target of the optimization of agent policy $\pi_\theta$ changes accordingly. 
\begin{remark}
At the early training stage, $\pi_{\widehat{\theta}}$ is of bad quality and represents the relatively bad part in unlabeled imperfect demonstrations. This makes the residual policy $\pi_\epsilon$ occupy the optimal mode within unlabeled demonstrations. Under such cases, agent policy $\pi_\theta$ is imitating the optimal demonstrations.
\label{remark1}
\end{remark}
\begin{theorem}
For the agent policy $\pi_\theta$ fixed, the optimal discriminator $D_\psi^\ast(s,a)$ can be written as
\begin{align}
    D_\psi^\ast(s,a)=\frac{\rho_{\pi_\epsilon}(s,a)}{\rho_{\pi_\epsilon}(s,a)+\frac{1-\alpha}{\alpha}\rho_{\pi_\theta}(s,a)},
\end{align}
With the optimal discriminator $D_\psi^\ast(s,a)$ fixed, the optimization of $\pi_\theta$ is equivalent to minimize
\begin{align}
    C + (1-\alpha) KL(\rho_{\pi_\epsilon}||\rho_{\pi_e}) + \alpha KL(\rho_{\pi_\theta}||\rho_{\pi_e}),
\end{align}
where $C=(1-\alpha)\log(1-\alpha) + \alpha\log\alpha$.
The global minimum of the proposed objective function is achieved if and only if $\rho_{\pi_\theta}=\rho_{\pi_\epsilon}=\rho_{\pi_e}$. At that point, the objective achieves the value $(1-\alpha)\log(1-\alpha) + \alpha\log\alpha$, and $D_\psi^\ast(s,a)$ achieves the value $\alpha$.
\label{theo2}
\end{theorem}

From Theorem \ref{theo2}, we prove that UID approaches Nash equilibrium when $\rho_{\pi_\theta}=\rho_{\pi_e}$. This illustrates that UID makes the agent imitate $\pi_e$ finally. Recall that we also show that $\pi_\theta$ is imitating the optimal demonstrations at the early training stage in Remark \ref{remark1}. Therefore, we conclude that UID makes $\pi_\theta$ imitate optimal demonstrations within unlabeled demonstrations firstly and then gradually covers more demonstrations in unlabeled imperfect demonstrations. This actually leads UID to relate to curriculum learning \cite{bengio2009curriculum} and self-paced learning \cite{kumar2010self}, which also make the model learn from good samples to other samples gradually. This connection provides a theoretical guarantee of UID's advantage compared to plain GAIL. The empirical study in the experiment part identifies the analysis above.

\subsection{Discussion}
\subsubsection{Connection with PU Learning}
The discriminator training scheme above is related to non-negative positive-unlabeled learning \cite{du2014analysis,kiryo2017positive,xu2017multi,xu2019positive}. In positive-unlabeled classification, two sets of data are sampled independently from positive data distribution $p_p(x)$ and unlabeled data distribution $p_u(x)$ as $\mathcal{X}_p=\{x_i^P\}_{i=1}^{n_p}\sim p_p(x)$ and $\mathcal{X}_u=\{x_i^U\}_{i=1}^{n_u}\sim p_u(x)$, and a classifier $g(x)$ needs to be trained to distinguish samples from $\mathcal{X}_p$ and $\mathcal{X}_u$. Regarding $\rho_{\pi_{\widehat{\theta}}}$ as the known positive distribution $p_p$ and $\rho_{\pi_e}$ as the unlabeled mixture data distribution $p_u$, we find that the process of discriminator training can be exactly viewed as a special example of PU learning. Moreover, we investigate the compatibility of PU learning with the adversarial imitation learning framework and show it can well handle imperfect demonstrations issue in adversarial imitation learning.  

Another related method is PU-GAIL, which also adopts a PU-based classifier in adversarial imitation learning~\cite{guo2020positive}. Under the assumption that the agent policy produces diverse demonstrations during training, PU-GAIL treats agent demonstrations as unlabeled data while regarding expert demonstrations as positive data to form this PU classifier.
PU-GAIL can be regarded as a regularization technology for the discriminator to prevent overfitting problems \cite{orsini2021matters}, which may help to stabilize the adversarial training. But PU-GAIL would fail when dealing with imperfect demonstrations, since it still lets the agent imitate all demonstrations equally all the time. By contrast, UID views expert demonstrations as unlabeled data and learns from the demonstrations via a self-paced way. Empirical results in the experiment show that UID has a better discriminative ability within unlabeled demonstrations and can achieve better performance with imperfect demonstrations.

\section{Experiments}
In this section, we conduct experiments to verify the effectiveness of UID in various benchmarks (\textit{i.e.}, MuJoCo \cite{todorov2012mujoco} and Robosuite \cite{robosuite2020}) under different settings. The experimental results demonstrate the advantage of UID from different aspects.\footnote{https://github.com/yunke-wang/UID} 

\subsubsection{Experimental Setting}
We evaluate UID on three MuJoCo \cite{todorov2012mujoco} locomotion tasks (\textit{i.e.}, Ant-v2, HalfCheetah-v2 and Walker2d-v2) firstly. 
The agent performance in MuJoCo can be measured by both the average cumulative rewards along trajectories and the final location of the agent (\textit{i.e.}, higher the better). We evaluate the agent every 5,000 transitions in training and the reported results are the average of the last 100 evaluations.  We repeat experiments for 5 trials with different random seeds.
To verify the robustness of UID with real-world human operation demonstrations, we also conduct experiments on a robot control task in Robosuite \cite{robosuite2020}. 

\subsubsection{Source of Demonstrations}
We collect a mixture of optimal and non-optimal demonstrations to conduct experiments. To form these unlabeled demonstrations, an optimal expert policy $\pi_o$ trained by TRPO is used to sample optimal demonstrations $\mathcal{D}_o$, and then 3 non-optimal expert policies $\pi_{n}$ are used to sample non-optimal demonstrations $\mathcal{D}_n$. 
Following existing works, we use two different kinds of $\pi_n$ to sample non-optimal demonstrations.
\begin{itemize}
    \item \textbf{D1}: We save 3 checkpoints during the RL training as 3 non-optimal expert policies $\pi_n$.
    \item \textbf{D2}: We add Gaussian noise $\xi$ to the action distribution $a^\ast$ of $\pi_o$ to form non-optimal expert $\pi_n$. The action of $\pi_n$ is modeled as $a\sim\mathcal{N}(a^\ast, \xi^2)$ and we choose $\xi=[0.25, 0.4, 0.6]$ in these 3 non-optimal policies.
\end{itemize}

Equal demonstrations are sampled from each policy. The unlabeled expert demonstrations $\mathcal{D}_e$ is formed by mixing the sampled optimal demonstrations $\mathcal{D}_o$ and non-optimal demonstrations $\mathcal{D}_n$. The data quality and the detailed implementation are deferred to the supplementary material.

\begin{figure}[!b]
	\centering
	\includegraphics[width=0.48\textwidth]{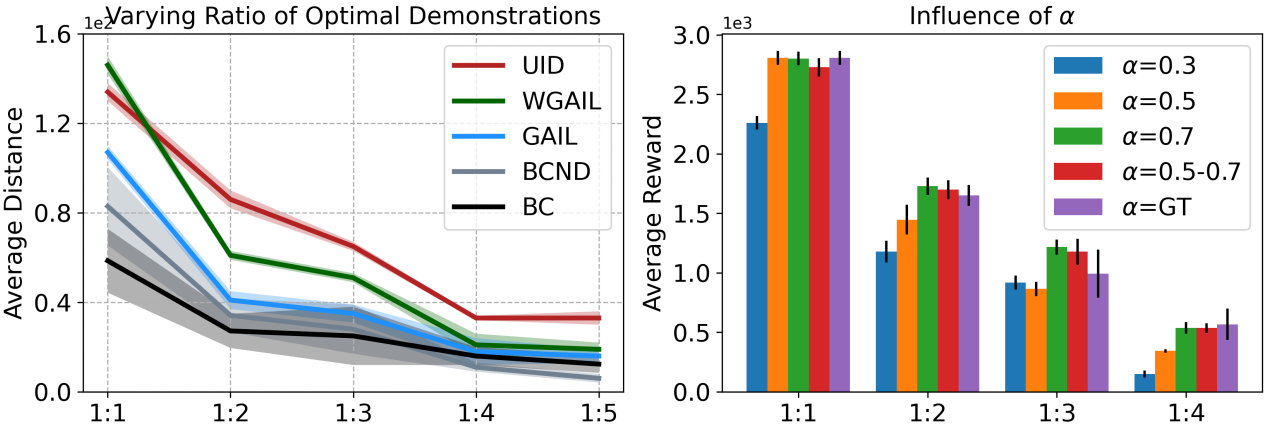}
	\caption{Performance with varying ratios of optimal demonstrations.}
	\label{fig_1}
\end{figure}

\begin{table*}[!t]  
	\centering 
	\resizebox{\textwidth}{!}{
    \begin{tabular}{lcccccc}
		\toprule  
		\multirow{2}{*}{Method}&  
		\multicolumn{3}{c}{D1}&\multicolumn{3}{c}{D2}\cr  
		\cmidrule(lr){2-4} \cmidrule(lr){5-7}
		&Ant-v2 &HalfCheetah-v2 &Walker2d-v2 &Ant-v2 &HalfCheetah-v2&Walker2d-v2\cr  
		\midrule 
	    WAIL \cite{xiao2019wasserstein}    & 1348$\pm$120 & 2282$\pm$58 & 2180$\pm$46   & 2039$\pm$48 & 3124$\pm$334  & 2656$\pm$170 \cr 
	    \rowcolor{mygray}
		UID-WAIL (Ours)  &\bf 1709$\pm$118  &\bf 2569$\pm$157 &\bf 2359$\pm$43 &\bf 2490$\pm$59  &\bf 3582$\pm$340 &\bf 3364$\pm$104  \cr  
		\midrule
    	GAIL \cite{ho2016generative}  & 1179$\pm$158 & 2159$\pm$139 & 1873$\pm$115 & 1797$\pm$137  & 2758$\pm$205 & 2786$\pm$262  \cr
    	\rowcolor{mygray}
		UID-GAIL (Ours) & \bf 1674$\pm$142 & \bf 3276$\pm$114 & \bf 2482$\pm$65 & \bf 2426$\pm$110  & \bf 3983$\pm$179 & \bf 3343$\pm$180   \cr  
		\midrule
		2IWIL \cite{wu2019imitation}  & 1591$\pm$71 & 2704$\pm$129 & 2204$\pm$66 &2317$\pm$123& 2656$\pm$261 & 2749$\pm$258 \cr
		\rowcolor{mygray}
    	IC-GAIL \cite{wu2019imitation}  &\bf  1974$\pm$41 & 2779$\pm$92  & 2002$\pm$54  & 1883$\pm$90 & 3087$\pm$226 & 2429$\pm$166   \cr
    	T-REX \cite{brown2019extrapolating} & -556$\pm$83 &2223$\pm$255 & 1866$\pm$296   & -22$\pm$2 & 1399$\pm$499 & 1622$\pm$165 \cr
    	\rowcolor{mygray}
		D-REX \cite{brown2020better} &  -1751$\pm$194  & 470$\pm$86  & 529$\pm$91  & -27$\pm$20  & 2588$\pm$75   &  1433$\pm$104   \cr
    	PU-GAIL \cite{xu2021positive}  &310$\pm$86 &1136$\pm$332 & 1469$\pm$379 & 1734$\pm$140&2413$\pm$505 & 2652$\pm$112   \cr
		\bottomrule
	\end{tabular}}
	\caption{Performance of proposed methods and compared methods in MuJoCo tasks with both stage 1 and stage 2 demonstrations, which is measured by the average and standard variance of ground-truth cumulative reward along 10 trajectories, \textit{i.e.}, higher average value is better. The value in \textbf{Bold} denotes the best value between UID and its baseline.
	} 
	\label{t2} 
\end{table*}

\subsection{Results on MuJoCo}
\subsubsection{Varying Ratios of Optimal Demonstrations}
We firstly investigate the capacity of UID when dealing with varying ratios of optimal demonstrations in Ant-v2 task. We begin with 50\% (1:1) optimal demonstrations, and gradually decrease the ratio of optimal data to around 16.7\% (1:5). The compared method are two state-of-the-art confidence-based methods WGAIL \cite{wang2021learning} and BCND \cite{sasaki2021behavioral} that do not require any prior information when estimating weight.

As claimed in \cite{sasaki2021behavioral}, BCND needs a ``50\% optimal data'' assumption on the mixed demonstrations to ensure a promising performance. If non-optimal demonstrations occupy the major mode, the confidence distribution is likely to drift to the non-optimal part.
We observe a similar phenomenon in our experiment. As shown in Figure \ref{fig_1}, when given 50\% optimal demonstrations, BCND can still outperform BC by a clear margin. However, when the ratio of optimal demonstrations decreases, the performance of BCND drops and starts to inferior to BC with less than 25\% optimal demonstrations. Online imitation learning methods (\textit{i.e.}, UID, WGAIL, and GAIL) perform generally better than offline imitation learning methods. WGAIL performs best at 50\% optimal demonstrations point, however, its performance decreases rapidly and achieves similar performance with GAIL when given less than 25\% optimal demonstrations point. By contrast, the curve of UID is clearly above GAIL as the data quality decreases. We therefore conclude that UID can have a better performance than WGAIL and BCND with limited optimal demonstrations. 

\subsubsection{Impact of $\alpha$}
We conduct ablation studies on $\alpha$ to find how different $\alpha$ could influence the final results of UID. We evaluate the performance of UID with varying ratios of optimal demonstrations with different $\alpha$ (\textit{i.e.}, $\alpha=0.3,0.5,0.7, 0.5-0.7$). We also provide a result by heuristically setting $\alpha$ as the real ratio of optimal demonstrations.
The results are summarized in Figure \ref{fig_1}. The 'red' rectangle denotes that we set $\alpha$ as the real ratio of non-optimal demonstrations. We find that UID enjoys a relatively considerable tolerance of $\alpha$. Generally, setting $\alpha=0.7$ results in the best performance in most cases. We therefore consider directly treating $\alpha$ as a hyper-parameter and UID can also be regarded as a method that does not require prior information.

\subsubsection{Performance on various AIL frameworks}
\begin{figure}[!t]
    \centering
    \includegraphics[width=0.47\textwidth]{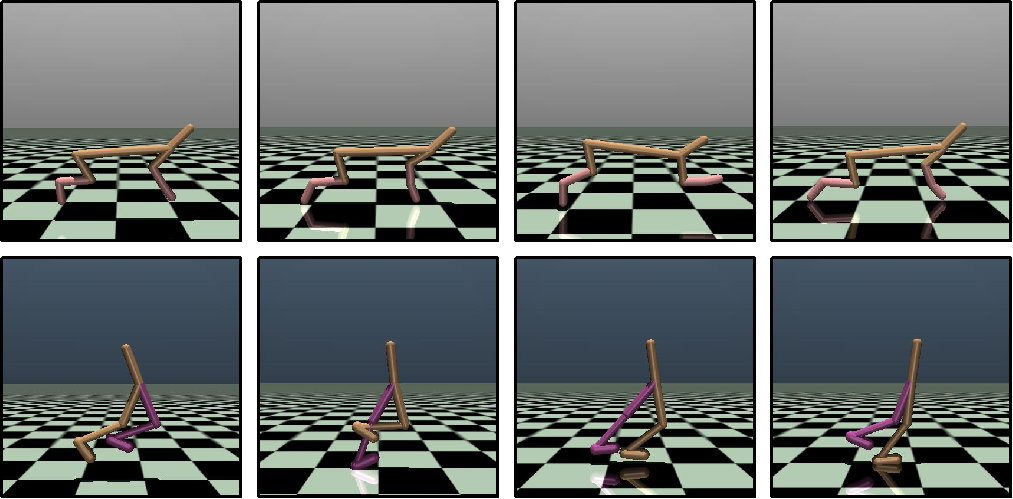}
    \caption{Visualization of the agent trained by UID with class 1 demonstrations. Time step increases from the leftmost figure (t=25) to the rightmost figure (t=100).}
    \label{learned_agent}
\end{figure}

Since UID can be extended into more adversarial imitation learning frameworks by defining different $f$-divergence in Theorem \ref{theo1}, we test the capacity of UID with two AIL baselines, \textit{i.e.}, GAIL and WAIL. The results are shown in Table \ref{t2}. 
We observe that UID beats vanilla AIL with both D1 and D2 demonstrations in all three baselines with a clear improvement. This illustrates the effectiveness of UID when dealing with different kinds of mixed imperfect demonstrations. We also conduct student's t-test on the results and the null hypothesis is the performance of UID is similar or worse than the GAIL baseline. The result is shown in Table \ref{ptest}, from which we can observe that there is a statistical significance between the performance of UID and GAIL since most p-values are clearly below 0.05.
We also provide screenshots in MuJoCo software to observe the performance of the agent from the visual perspective, as shown in Figure \ref{learned_agent}. We find that the agent learned by UID runs fast and can be successfully qualified for these tasks. 
Additionally, we compare UID with several preference-based methods (\textit{i.e.}, T-REX and D-REX) and confidence-based methods (\textit{i.e.}, 2IWIL and IC-GAIL). 
The rankings of trajectories in T-REX are given as a prior and we use the normalized reward of each checkpoint as the confidence for each demonstration. 
Generally speaking, preference-based methods do not perform well in most cases, yet the confidence-based methods 2IWIL and IC-GAIL perform clearly better. Especially in Ant-v2 and Walker-v2, we find that the performance of 2IWIL in these two environments is only slightly inferior to UID. However, 2IWIL requires strong prior information on the confidence of demonstrations that may not be easily obtained. 

\begin{table}[!b]  
	\centering 
    \begin{tabular}{lccc}
		\toprule 
		p-value & Ant-v2 & HalfCheetah-v2 & Walker2d-v2 \cr  
		\midrule
		(D1) & 0.0702 & 0.0005 & 0.0032  \cr 
	    (D2) & 0.0126 & 0.0038 & 0.1556  \cr 
		\bottomrule
	\end{tabular}
	\caption{The p-value between UID and its baseline GAIL.} 
	\label{ptest} 
\end{table}

\begin{figure*}[!t]
    \centering
    \subfigure{
    \begin{minipage}[t]{0.32\linewidth}
        \centering
        \includegraphics[width=2.3in]{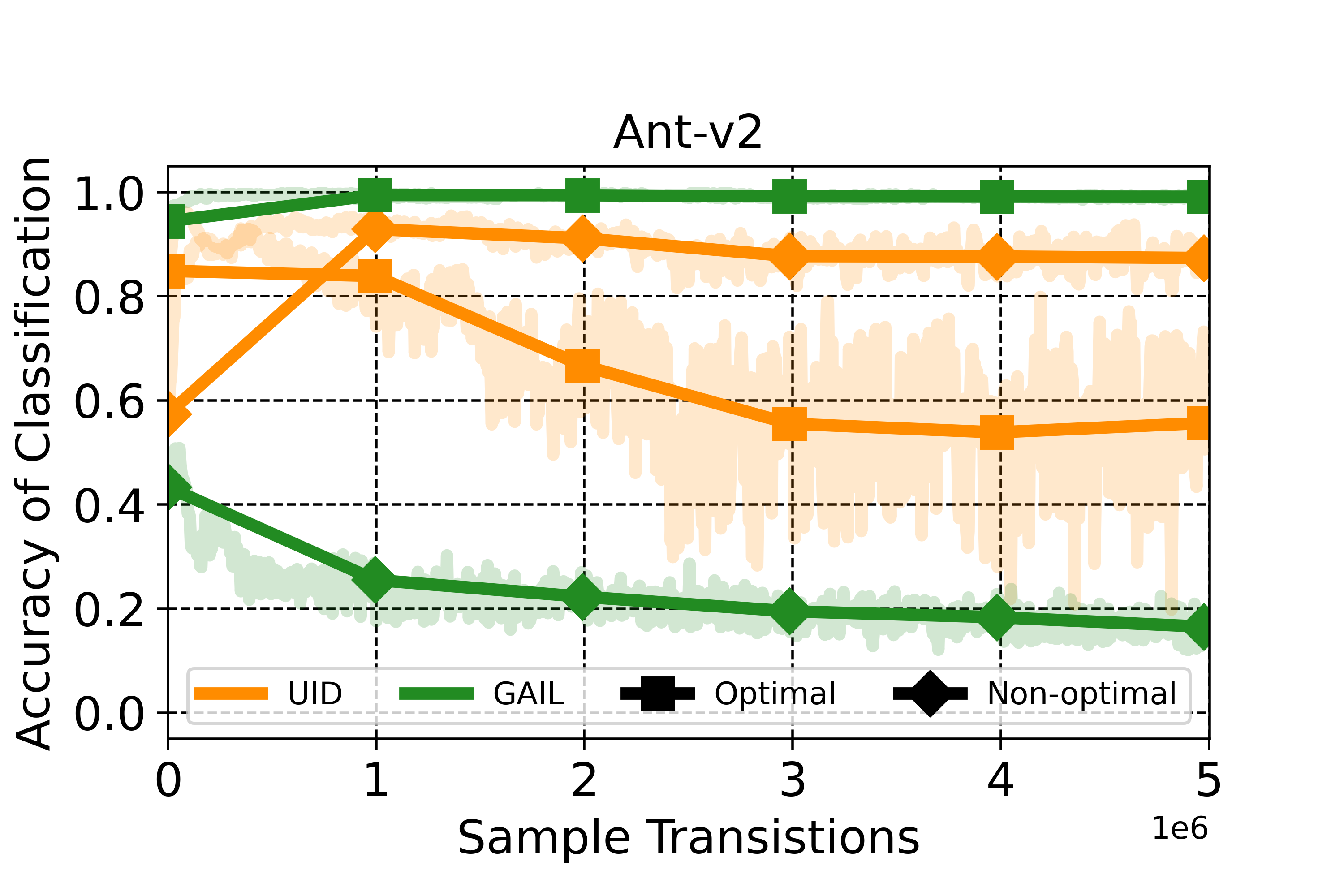}
    \end{minipage}}
    \subfigure{
    \begin{minipage}[t]{0.32\linewidth}
        \centering
        \includegraphics[width=2.3in]{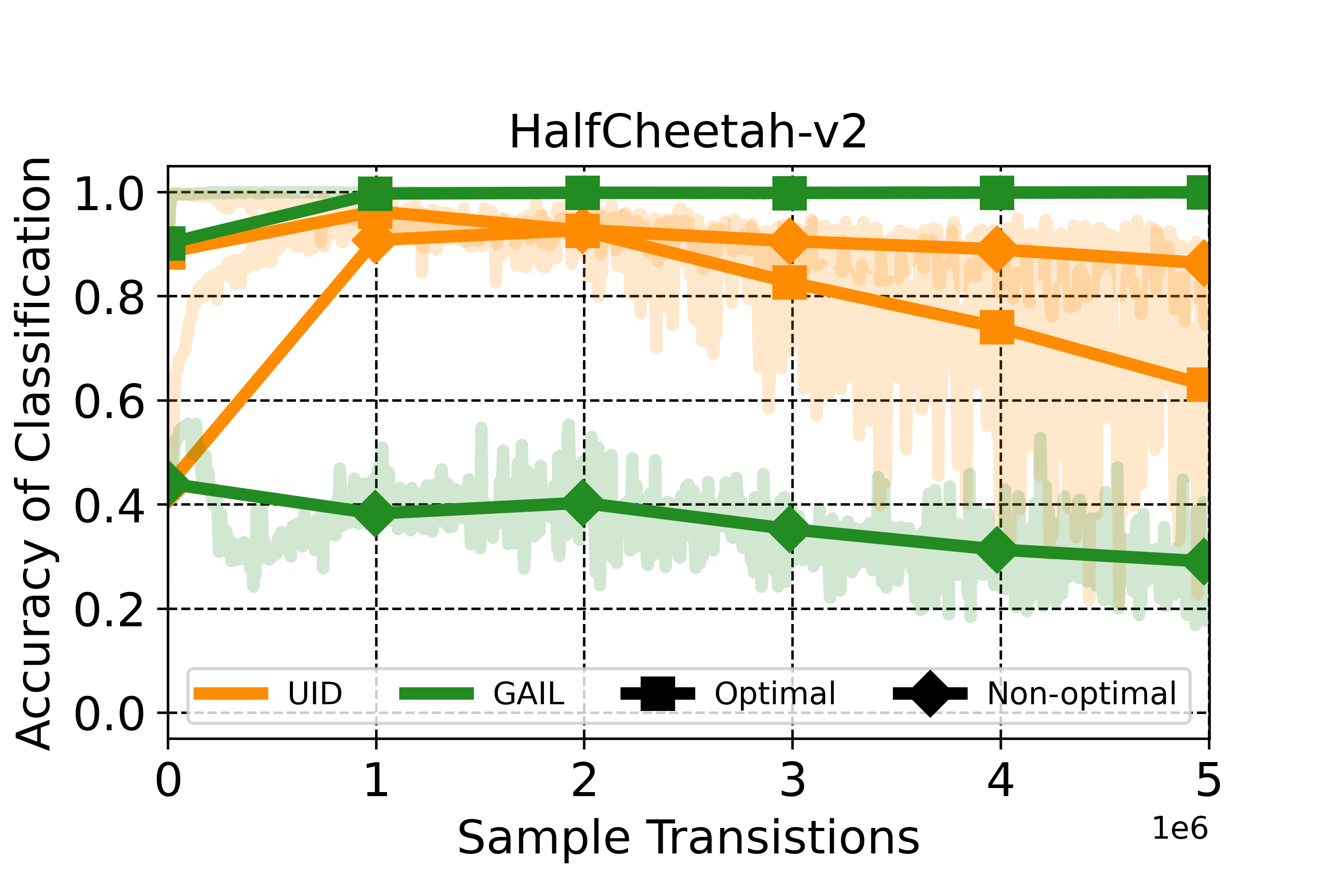}
    \end{minipage}}
    \subfigure{
    \begin{minipage}[t]{0.32\linewidth}
        \centering
        \includegraphics[width=2.3in]{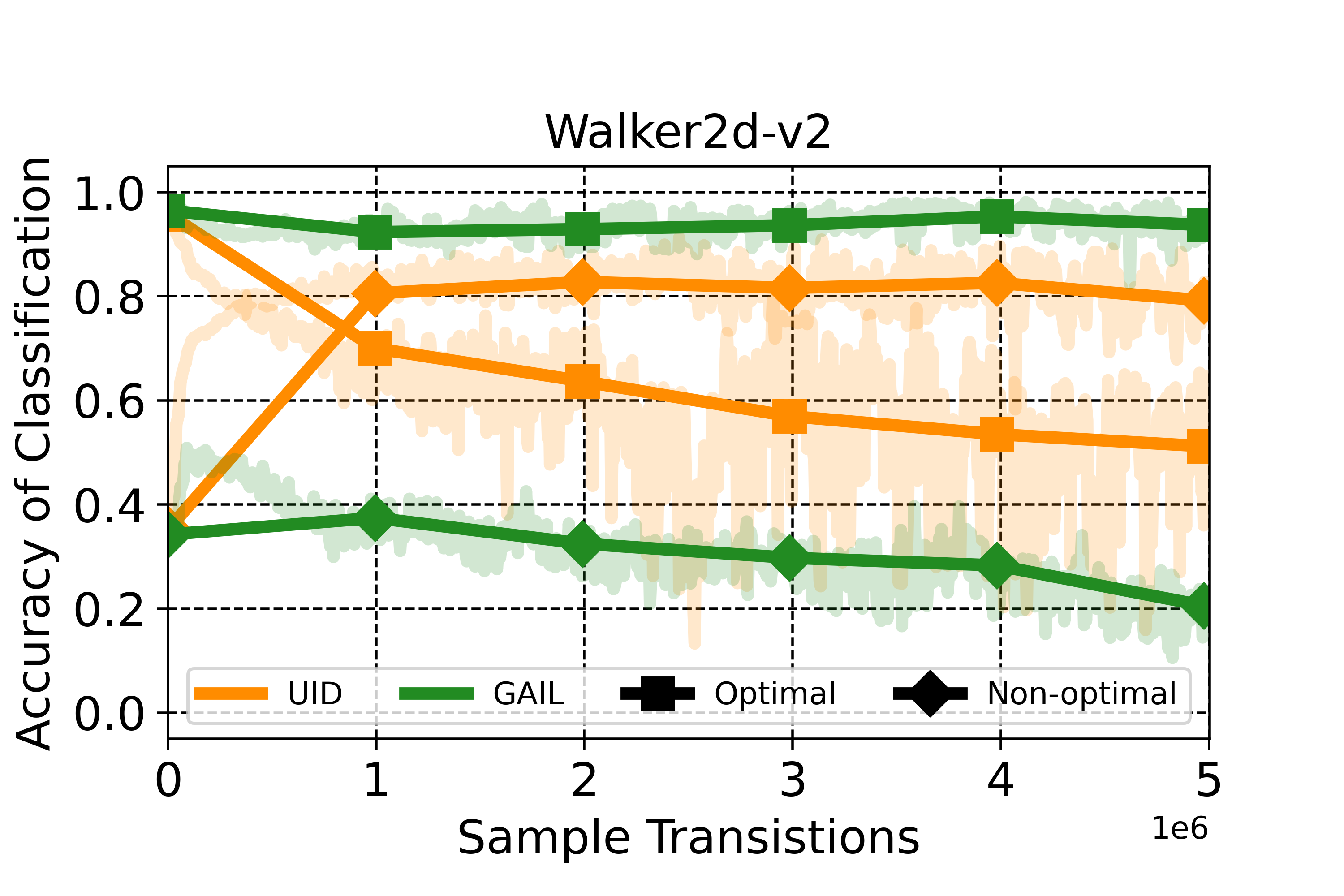}
    \end{minipage}}
    \caption{The accuracy of $D_\psi$ in classifying optimal demonstrations $\mathcal{D}_o$ and non-optimal demonstrations $\mathcal{D}_n$ within unlabeled demonstrations $\mathcal{D}_e$ during UID and GAIL training. We provide smooth version of the initial learning curve (the shade part) for better observation.}
    \label{fig_d}
\end{figure*}

As discussed in the methodology, PU-GAIL also introduces a PU classifier into a generative adversarial imitation learning framework. While treating agent demonstrations as unlabeled samples, PU-GAIL learns a better discriminator by considering the increasing ratio of good samples produced by agent policy. This training scheme is more sound than GAIL training and might be helpful to stable GAN training and avoid local minima, however, this does not change its actual upper bound of performance since PU-GAIL still regards all given expert demonstrations as positives. When given imperfect demonstrations, PU-GAIL can only learn an inferior performance.
In Table \ref{t2}, we observe the performance of PU-GAIL is similar or sometimes inferior to its baseline. This illustrates that PU-GAIL can not well handle imperfect demonstrations in imitation learning.

\begin{figure}[!b]
	\centering
	\includegraphics[width=0.47\textwidth]{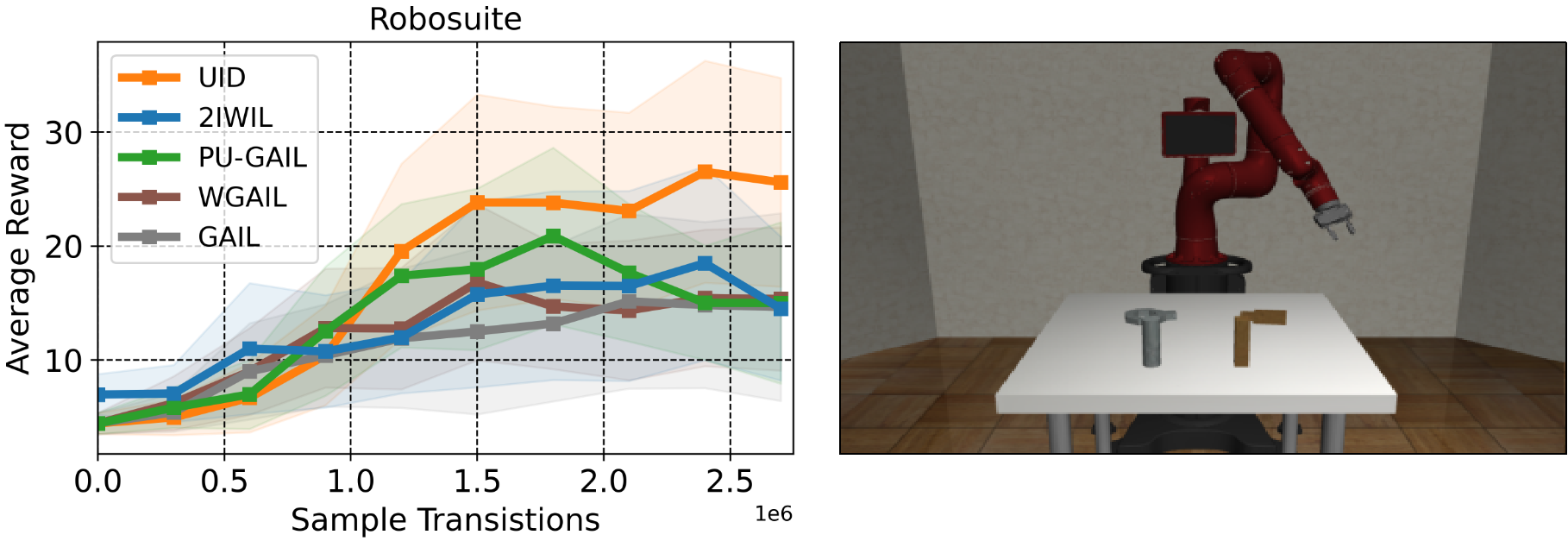}
	\caption{Performance of UID in RoboSuite tasks.}
	\label{fig_robo}
\end{figure}
\subsubsection{Analysis on the discriminator}
During the training of UID on unlabeled demonstrations $\mathcal{D}_e$, we investigate the performance of the discriminator by testing its classification accuracy on optimal demonstrations $\mathcal{D}_o$ and non-optimal demonstrations $\mathcal{D}_n$. The accurate classification is defined as treating $\mathcal{D}_o$ as positive and treating $\mathcal{D}_n$ as negative.
In Figure \ref{fig_d}, there is a clear trend that the discriminator in both methods reaches high accuracy in classifying $\mathcal{D}_o$. However, when it comes to $\mathcal{D}_n$, the accuracy of the discriminator is generally low in GAIL. This shows that the discriminator in GAIL equally regards $\mathcal{D}_n$ and $\mathcal{D}_o$ as `positive', while the discriminator in UID obviously has discriminative ability on these two kinds of demonstrations within unlabeled demonstrations $\mathcal{D}_e$. This is due to introducing the idea of PU classification in UID.

Another trend is that the discriminator in UID has a decreased classification accuracy on optimal demonstrations $\mathcal{D}_o$ during training.
Since the output of the discriminator is proportional to the reward, agent demonstrations that are classified as `positive' by the discriminator will be assigned with a high reward in the RL step. 
At the beginning of the training, the obtained reward of agent demonstrations that resemble $\mathcal{D}_o$ would be thus relatively higher. This can be exactly viewed as encouraging the agent to learn from $\mathcal{D}_o$ at first. As the training progresses, the reward of agent demonstrations that are close to $\mathcal{D}_o$ will decline accordingly. This enables a chance for those non-optimal demonstrations $\mathcal{D}_u$ to participate in and guide the agent training. The empirical results here identify our analysis on the connection between UID and self-paced learning. 

\subsection{Results on RoboSuite Platform}
We also evaluate the robustness of UID on the RoboSuite platform \cite{robosuite2020} with real-world demonstrations. We consider a ``Nut Assembly'' task in Saywer, in which two colored pegs and two colored nuts are mounted on the tabletop, as shown in the right of Figure \ref{fig_robo}. The robot aims to fit the nut into its related peg. We use real-world demonstrations by human operators from RoboTurk website\footnote{https://roboturk.stanford.edu/dataset\_sim.html}. The demonstrations contain 10 trajectories with approaching length and the overall number of demonstrations is 5000. Based on the accumulative reward of trajectories, only three trajectories are regarded as optimal demonstrations. We therefore expect to test the performance of UID in imperfect demonstrations from the real world. Figure \ref{fig_robo} shows the performance of UID with 3 million transition samples for RL training. We find that UID performs best over the other 4 compared methods. This experiment further identifies the robustness of UID with human demonstrations. 

\section{Conclusion}
In this paper, we propose a general framework called UID to address the unlabeled imperfect demonstrations problem in adversarial imitation learning. Instead of treating all imperfect demonstrations as absolutely positive in plain GAIL, we regard imperfect demonstrations as unlabeled data and adopt a more efficient scheme to make the agent learn from them. With a fraction of unlabeled demonstrations separated to match the agent demonstrations, we develop a positive-unlabeled adversarial imitation learning framework. 
We also make this technology compatible with various adversarial imitation learning baselines. The final experimental results on MuJoCo and RoboSuite platforms demonstrate the advantage of UID in dealing with imperfect demonstrations over other compared methods. 

\section{Acknowledgements}
This work was supported in part by the National Natural Science Foundation of China under Grants 62225113, 41871243, 62141112, the Science and Technology Major Project of Hubei Province (Next-Generation AI Technologies) under Grant 2019AEA170, the Australian Research Council under Project DP210101859, and the University of Sydney Research Accelerator (SOAR) Prize.

\bibliography{aaai23}

\begin{thebibliography}{56}
\providecommand{\natexlab}[1]{#1}

\bibitem[{Amodei et~al.(2016)Amodei, Olah, Steinhardt, Christiano, Schulman,
  and Man{\'e}}]{amodei2016concrete}
Amodei, D.; Olah, C.; Steinhardt, J.; Christiano, P.; Schulman, J.; and
  Man{\'e}, D. 2016.
\newblock Concrete problems in AI safety.
\newblock \emph{arXiv preprint arXiv:1606.06565}.

\bibitem[{Bengio et~al.(2009)Bengio, Louradour, Collobert, and
  Weston}]{bengio2009curriculum}
Bengio, Y.; Louradour, J.; Collobert, R.; and Weston, J. 2009.
\newblock Curriculum learning.
\newblock In \emph{Proceedings of the 26th annual international conference on
  machine learning}, 41--48.

\bibitem[{Brantley, Sun, and Henaff(2019)}]{brantley2019disagreement}
Brantley, K.; Sun, W.; and Henaff, M. 2019.
\newblock Disagreement-regularized imitation learning.
\newblock In \emph{International Conference on Learning Representations}.

\bibitem[{Brown et~al.(2019)Brown, Goo, Nagarajan, and
  Niekum}]{brown2019extrapolating}
Brown, D.; Goo, W.; Nagarajan, P.; and Niekum, S. 2019.
\newblock Extrapolating beyond suboptimal demonstrations via inverse
  reinforcement learning from observations.
\newblock In \emph{International conference on machine learning}, 783--792.
  PMLR.

\bibitem[{Brown, Goo, and Niekum(2020)}]{brown2020better}
Brown, D.~S.; Goo, W.; and Niekum, S. 2020.
\newblock Better-than-Demonstrator Imitation Learning via Automatically-Ranked
  Demonstrations.
\newblock In \emph{Conference on Robot Learning}, 330--359.

\bibitem[{Cai et~al.(2021)Cai, Ding, Chen, Jiang, Sugiyama, and
  Zhou}]{cai2021seeing}
Cai, X.-Q.; Ding, Y.-X.; Chen, Z.-X.; Jiang, Y.; Sugiyama, M.; and Zhou, Z.-H.
  2021.
\newblock Seeing Differently, Acting Similarly: Heterogeneously Observable
  Imitation Learning.
\newblock \emph{arXiv preprint arXiv:2106.09256}.

\bibitem[{Chen, Paleja, and Gombolay(2021)}]{chen2021learning}
Chen, L.; Paleja, R.; and Gombolay, M. 2021.
\newblock Learning from Suboptimal Demonstration via Self-Supervised Reward
  Regression.
\newblock In \emph{Conference on Robot Learning}, 1262--1277. PMLR.

\bibitem[{Chen et~al.(2022)Chen, Cai, Jiang, and Zhou}]{chen2022anomaly}
Chen, Z.-X.; Cai, X.-Q.; Jiang, Y.; and Zhou, Z.-H. 2022.
\newblock Anomaly Guided Policy Learning from Imperfect Demonstrations.
\newblock In \emph{Proceedings of the 21st International Conference on
  Autonomous Agents and Multiagent Systems}, 244--252.

\bibitem[{Christiano et~al.(2017)Christiano, Leike, Brown, Martic, Legg, and
  Amodei}]{christiano2017deep}
Christiano, P.~F.; Leike, J.; Brown, T.; Martic, M.; Legg, S.; and Amodei, D.
  2017.
\newblock Deep reinforcement learning from human preferences.
\newblock In \emph{Advances in Neural Information Processing Systems},
  4299--4307.

\bibitem[{Dadashi et~al.(2020)Dadashi, Hussenot, Geist, and
  Pietquin}]{dadashi2020primal}
Dadashi, R.; Hussenot, L.; Geist, M.; and Pietquin, O. 2020.
\newblock Primal Wasserstein Imitation Learning.
\newblock In \emph{International Conference on Learning Representations}.

\bibitem[{Du~Plessis, Niu, and Sugiyama(2014)}]{du2014analysis}
Du~Plessis, M.~C.; Niu, G.; and Sugiyama, M. 2014.
\newblock Analysis of learning from positive and unlabeled data.
\newblock \emph{Advances in neural information processing systems}, 27.

\bibitem[{Fan et~al.(2018)Fan, Zhu, Zhu, Liu, Zeng, Gupta, Creus-Costa,
  Savarese, and Fei-Fei}]{fan2018surreal}
Fan, L.; Zhu, Y.; Zhu, J.; Liu, Z.; Zeng, O.; Gupta, A.; Creus-Costa, J.;
  Savarese, S.; and Fei-Fei, L. 2018.
\newblock Surreal: Open-source reinforcement learning framework and robot
  manipulation benchmark.
\newblock In \emph{Conference on Robot Learning}, 767--782. PMLR.

\bibitem[{Fu, Luo, and Levine(2018)}]{fu2018learning}
Fu, J.; Luo, K.; and Levine, S. 2018.
\newblock Learning Robust Rewards with Adverserial Inverse Reinforcement
  Learning.
\newblock In \emph{International Conference on Learning Representations}.

\bibitem[{Goodfellow et~al.(2014)Goodfellow, Pouget-Abadie, Mirza, Xu,
  Warde-Farley, Ozair, Courville, and Bengio}]{goodfellow2014generative}
Goodfellow, I.; Pouget-Abadie, J.; Mirza, M.; Xu, B.; Warde-Farley, D.; Ozair,
  S.; Courville, A.; and Bengio, Y. 2014.
\newblock Generative adversarial nets.
\newblock In \emph{Advances in neural information processing systems},
  2672--2680.

\bibitem[{Guo et~al.(2020)Guo, Xu, Huang, Wang, Shi, Xu, and
  Tao}]{guo2020positive}
Guo, T.; Xu, C.; Huang, J.; Wang, Y.; Shi, B.; Xu, C.; and Tao, D. 2020.
\newblock On positive-unlabeled classification in GAN.
\newblock In \emph{Proceedings of the IEEE/CVF Conference on Computer Vision
  and Pattern Recognition}, 8385--8393.

\bibitem[{Haarnoja et~al.(2018)Haarnoja, Zhou, Abbeel, and
  Levine}]{haarnoja2018soft}
Haarnoja, T.; Zhou, A.; Abbeel, P.; and Levine, S. 2018.
\newblock Soft actor-critic: Off-policy maximum entropy deep reinforcement
  learning with a stochastic actor.
\newblock In \emph{International conference on machine learning}, 1861--1870.
  PMLR.

\bibitem[{Ho and Ermon(2016)}]{ho2016generative}
Ho, J.; and Ermon, S. 2016.
\newblock Generative adversarial imitation learning.
\newblock In \emph{Advances in neural information processing systems},
  4565--4573.

\bibitem[{Hussein et~al.(2017)Hussein, Gaber, Elyan, and
  Jayne}]{hussein2017imitation}
Hussein, A.; Gaber, M.~M.; Elyan, E.; and Jayne, C. 2017.
\newblock Imitation learning: A survey of learning methods.
\newblock \emph{ACM Computing Surveys (CSUR)}, 50(2): 1--35.

\bibitem[{Kaelbling, Littman, and Moore(1996)}]{kaelbling1996reinforcement}
Kaelbling, L.~P.; Littman, M.~L.; and Moore, A.~W. 1996.
\newblock Reinforcement learning: A survey.
\newblock \emph{Journal of artificial intelligence research}, 4: 237--285.

\bibitem[{Ke et~al.(2020)Ke, Choudhury, Barnes, Sun, Lee, and
  Srinivasa}]{ke2020imitation}
Ke, L.; Choudhury, S.; Barnes, M.; Sun, W.; Lee, G.; and Srinivasa, S. 2020.
\newblock Imitation learning as f-divergence minimization.
\newblock In \emph{International Workshop on the Algorithmic Foundations of
  Robotics}, 313--329. Springer.

\bibitem[{Kim et~al.(2021)Kim, Seo, Lee, Jeon, Hwang, Yang, and
  Kim}]{kim2021demodice}
Kim, G.-H.; Seo, S.; Lee, J.; Jeon, W.; Hwang, H.; Yang, H.; and Kim, K.-E.
  2021.
\newblock DemoDICE: Offline Imitation Learning with Supplementary Imperfect
  Demonstrations.
\newblock In \emph{International Conference on Learning Representations}.

\bibitem[{Kiryo et~al.(2017)Kiryo, Niu, Du~Plessis, and
  Sugiyama}]{kiryo2017positive}
Kiryo, R.; Niu, G.; Du~Plessis, M.~C.; and Sugiyama, M. 2017.
\newblock Positive-unlabeled learning with non-negative risk estimator.
\newblock \emph{Advances in neural information processing systems}, 30.

\bibitem[{Kumar, Packer, and Koller(2010)}]{kumar2010self}
Kumar, M.~P.; Packer, B.; and Koller, D. 2010.
\newblock Self-paced learning for latent variable models.
\newblock In \emph{Advances in neural information processing systems},
  1189--1197.

\bibitem[{Li, Song, and Ermon(2017)}]{li2017infogail}
Li, Y.; Song, J.; and Ermon, S. 2017.
\newblock Infogail: Interpretable imitation learning from visual
  demonstrations.
\newblock In \emph{Advances in Neural Information Processing Systems},
  3812--3822.

\bibitem[{Liu et~al.(2022)Liu, Tang, Li, and Luo}]{liu2022robust}
Liu, L.; Tang, Z.; Li, L.; and Luo, D. 2022.
\newblock Robust Imitation Learning from Corrupted Demonstrations.
\newblock \emph{arXiv preprint arXiv:2201.12594}.

\bibitem[{Niu et~al.(2016)Niu, du~Plessis, Sakai, Ma, and
  Sugiyama}]{niu2016theoretical}
Niu, G.; du~Plessis, M.~C.; Sakai, T.; Ma, Y.; and Sugiyama, M. 2016.
\newblock Theoretical comparisons of positive-unlabeled learning against
  positive-negative learning.
\newblock \emph{Advances in neural information processing systems}, 29.

\bibitem[{Orsini et~al.(2021)Orsini, Raichuk, Hussenot, Vincent, Dadashi,
  Girgin, Geist, Bachem, Pietquin, and Andrychowicz}]{orsini2021matters}
Orsini, M.; Raichuk, A.; Hussenot, L.; Vincent, D.; Dadashi, R.; Girgin, S.;
  Geist, M.; Bachem, O.; Pietquin, O.; and Andrychowicz, M. 2021.
\newblock What matters for adversarial imitation learning?
\newblock \emph{Advances in Neural Information Processing Systems}, 34.

\bibitem[{Peng et~al.(2018)Peng, Kanazawa, Toyer, Abbeel, and
  Levine}]{peng2018variational}
Peng, X.~B.; Kanazawa, A.; Toyer, S.; Abbeel, P.; and Levine, S. 2018.
\newblock Variational Discriminator Bottleneck: Improving Imitation Learning,
  Inverse RL, and GANs by Constraining Information Flow.
\newblock In \emph{International Conference on Learning Representations}.

\bibitem[{Pomerleau(1988)}]{pomerleau1988alvinn}
Pomerleau, D.~A. 1988.
\newblock Alvinn: An autonomous land vehicle in a neural network.
\newblock \emph{Advances in neural information processing systems}, 1.

\bibitem[{Puterman(1994)}]{DBLP:books/wi/Puterman94}
Puterman, M.~L. 1994.
\newblock \emph{Markov Decision Processes: Discrete Stochastic Dynamic
  Programming}.
\newblock Wiley Series in Probability and Statistics. Wiley.
\newblock ISBN 978-0-47161977-2.

\bibitem[{Qiu et~al.(2022)Qiu, Yang, Wang, and Fu}]{qiu2022dynamic}
Qiu, Z.; Yang, Q.; Wang, J.; and Fu, D. 2022.
\newblock Dynamic Graph Reasoning for Multi-person 3D Pose Estimation.
\newblock In \emph{Proceedings of the 30th ACM International Conference on
  Multimedia}, 3521--3529.

\bibitem[{Ren, Yeh, and Schwing(2020)}]{ren2020not}
Ren, Z.; Yeh, R.; and Schwing, A. 2020.
\newblock Not all unlabeled data are equal: Learning to weight data in
  semi-supervised learning.
\newblock \emph{Advances in Neural Information Processing Systems}, 33:
  21786--21797.

\bibitem[{Ross, Gordon, and Bagnell(2011)}]{ross2011reduction}
Ross, S.; Gordon, G.; and Bagnell, D. 2011.
\newblock A reduction of imitation learning and structured prediction to
  no-regret online learning.
\newblock In \emph{Proceedings of the fourteenth international conference on
  artificial intelligence and statistics}, 627--635.

\bibitem[{Sasaki and Yamashina(2021)}]{sasaki2021behavioral}
Sasaki, F.; and Yamashina, R. 2021.
\newblock Behavioral Cloning from Noisy Demonstrations.
\newblock In \emph{International Conference on Learning Representations}.

\bibitem[{Schulman et~al.(2015)Schulman, Levine, Abbeel, Jordan, and
  Moritz}]{schulman2015trust}
Schulman, J.; Levine, S.; Abbeel, P.; Jordan, M.; and Moritz, P. 2015.
\newblock Trust region policy optimization.
\newblock In \emph{International conference on machine learning}, 1889--1897.

\bibitem[{Schulman et~al.(2017)Schulman, Wolski, Dhariwal, Radford, and
  Klimov}]{schulman2017proximal}
Schulman, J.; Wolski, F.; Dhariwal, P.; Radford, A.; and Klimov, O. 2017.
\newblock Proximal policy optimization algorithms.
\newblock \emph{arXiv preprint arXiv:1707.06347}.

\bibitem[{Silver et~al.(2016)Silver, Huang, Maddison, Guez, Sifre, Van
  Den~Driessche, Schrittwieser, Antonoglou, Panneershelvam, Lanctot
  et~al.}]{silver2016mastering}
Silver, D.; Huang, A.; Maddison, C.~J.; Guez, A.; Sifre, L.; Van Den~Driessche,
  G.; Schrittwieser, J.; Antonoglou, I.; Panneershelvam, V.; Lanctot, M.;
  et~al. 2016.
\newblock Mastering the game of Go with deep neural networks and tree search.
\newblock \emph{nature}, 529(7587): 484--489.

\bibitem[{Sutton and Barto(2018)}]{sutton2018reinforcement}
Sutton, R.~S.; and Barto, A.~G. 2018.
\newblock \emph{Reinforcement learning: An introduction}.
\newblock MIT press.

\bibitem[{Todorov, Erez, and Tassa(2012)}]{todorov2012mujoco}
Todorov, E.; Erez, T.; and Tassa, Y. 2012.
\newblock Mujoco: A physics engine for model-based control.
\newblock In \emph{2012 IEEE/RSJ International Conference on Intelligent Robots
  and Systems}, 5026--5033. IEEE.

\bibitem[{Tu et~al.(2022)Tu, Robey, Zhang, and Matni}]{tu2022sample}
Tu, S.; Robey, A.; Zhang, T.; and Matni, N. 2022.
\newblock On the sample complexity of stability constrained imitation learning.
\newblock In \emph{Learning for Dynamics and Control Conference}, 180--191.
  PMLR.

\bibitem[{Van~Hasselt, Guez, and Silver(2016)}]{van2016deep}
Van~Hasselt, H.; Guez, A.; and Silver, D. 2016.
\newblock Deep reinforcement learning with double q-learning.
\newblock In \emph{Proceedings of the AAAI conference on artificial
  intelligence}, volume~30.

\bibitem[{Wang, Xu, and Du(2021)}]{ijcai2021-0434}
Wang, Y.; Xu, C.; and Du, B. 2021.
\newblock Robust Adversarial Imitation Learning via Adaptively-Selected
  Demonstrations.
\newblock In \emph{Proceedings of the Thirtieth International Joint Conference
  on Artificial Intelligence}, 3155--3161.

\bibitem[{Wang et~al.(2021)Wang, Xu, Du, and Lee}]{wang2021learning}
Wang, Y.; Xu, C.; Du, B.; and Lee, H. 2021.
\newblock Learning to Weight Imperfect Demonstrations.
\newblock In \emph{International Conference on Machine Learning}, 10961--10970.
  PMLR.

\bibitem[{Wu et~al.(2019)Wu, Charoenphakdee, Bao, Tangkaratt, and
  Sugiyama}]{wu2019imitation}
Wu, Y.-H.; Charoenphakdee, N.; Bao, H.; Tangkaratt, V.; and Sugiyama, M. 2019.
\newblock Imitation learning from imperfect demonstration.
\newblock In \emph{International Conference on Machine Learning}, 6818--6827.
  PMLR.

\bibitem[{Xiao et~al.(2019)Xiao, Herman, Wagner, Ziesche, Etesami, and
  Linh}]{xiao2019wasserstein}
Xiao, H.; Herman, M.; Wagner, J.; Ziesche, S.; Etesami, J.; and Linh, T.~H.
  2019.
\newblock Wasserstein adversarial imitation learning.
\newblock \emph{arXiv preprint arXiv:1906.08113}.

\bibitem[{Xu and Denil(2021)}]{xu2021positive}
Xu, D.; and Denil, M. 2021.
\newblock Positive-Unlabeled Reward Learning.
\newblock In \emph{Conference on Robot Learning}, 205--219. PMLR.

\bibitem[{Xu et~al.(2022)Xu, Zhan, Yin, and Qin}]{xu2022discriminator}
Xu, H.; Zhan, X.; Yin, H.; and Qin, H. 2022.
\newblock Discriminator-weighted offline imitation learning from suboptimal
  demonstrations.
\newblock In \emph{International Conference on Machine Learning}, 24725--24742.
  PMLR.

\bibitem[{Xu, Li, and Yu(2020)}]{xu2020error}
Xu, T.; Li, Z.; and Yu, Y. 2020.
\newblock Error bounds of imitating policies and environments.
\newblock \emph{Advances in Neural Information Processing Systems}, 33:
  15737--15749.

\bibitem[{Xu et~al.(2019)Xu, Wang, Chen, Han, Xu, Tao, and Xu}]{xu2019positive}
Xu, Y.; Wang, Y.; Chen, H.; Han, K.; Xu, C.; Tao, D.; and Xu, C. 2019.
\newblock Positive-unlabeled compression on the cloud.
\newblock \emph{Advances in Neural Information Processing Systems}, 32.

\bibitem[{Xu et~al.(2017)Xu, Xu, Xu, and Tao}]{xu2017multi}
Xu, Y.; Xu, C.; Xu, C.; and Tao, D. 2017.
\newblock Multi-Positive and Unlabeled Learning.
\newblock In \emph{IJCAI}, 3182--3188.

\bibitem[{Zha et~al.(2021)Zha, Xie, Ma, Zhang, Lian, Hu, and
  Liu}]{zha2021douzero}
Zha, D.; Xie, J.; Ma, W.; Zhang, S.; Lian, X.; Hu, X.; and Liu, J. 2021.
\newblock Douzero: Mastering doudizhu with self-play deep reinforcement
  learning.
\newblock In \emph{International Conference on Machine Learning}, 12333--12344.
  PMLR.

\bibitem[{Zhang et~al.(2020)Zhang, Zhu, Niu, Han, Sugiyama, and
  Kankanhalli}]{zhang2020geometry}
Zhang, J.; Zhu, J.; Niu, G.; Han, B.; Sugiyama, M.; and Kankanhalli, M. 2020.
\newblock Geometry-aware Instance-reweighted Adversarial Training.
\newblock In \emph{International Conference on Learning Representations}.

\bibitem[{Zhang et~al.(2021)Zhang, Cao, Sadigh, and Sui}]{zhang2021confidence}
Zhang, S.; Cao, Z.; Sadigh, D.; and Sui, Y. 2021.
\newblock Confidence-Aware Imitation Learning from Demonstrations with Varying
  Optimality.
\newblock \emph{Advances in Neural Information Processing Systems}, 34:
  12340--12350.

\bibitem[{Zheng et~al.(2022)Zheng, Verma, Zhou, Tsang, and
  Chen}]{zheng2022imitation}
Zheng, B.; Verma, S.; Zhou, J.; Tsang, I.~W.; and Chen, F. 2022.
\newblock Imitation learning: Progress, taxonomies and challenges.
\newblock \emph{IEEE Transactions on Neural Networks and Learning Systems},
  1--16.

\bibitem[{Zhong, Du, and Xu(2021)}]{zhong2021learning}
Zhong, Y.; Du, B.; and Xu, C. 2021.
\newblock Learning to reweight examples in multi-label classification.
\newblock \emph{Neural Networks}, 142: 428--436.

\bibitem[{Zhu et~al.(2020)Zhu, Wong, Mandlekar, and
  Mart\'{i}n-Mart\'{i}n}]{robosuite2020}
Zhu, Y.; Wong, J.; Mandlekar, A.; and Mart\'{i}n-Mart\'{i}n, R. 2020.
\newblock robosuite: A Modular Simulation Framework and Benchmark for Robot
  Learning.
\newblock In \emph{arXiv preprint arXiv:2009.12293}.

\end{thebibliography}
\end{document}